\documentclass[10pt,twocolumn,letterpaper]{article}

\usepackage{cvpr}              

\definecolor{cvprblue}{rgb}{0.21,0.49,0.74}
\usepackage[pagebackref,breaklinks,colorlinks,allcolors=cvprblue]{hyperref}

\usepackage{multirow} 
\usepackage{float} 

\def\paperID{773} 
\def\confName{CVPR}
\def\confYear{2025}

\title{Degradation-Aware Feature Perturbation for All-in-One Image Restoration}

\author{Xiangpeng Tian, Xiangyu Liao, Xiao Liu, Meng Li, Chao Ren\thanks{Corresponding author}\\
College of Electronics and Information Engineering, Sichuan University, China\\
{\tt\small \{tianxp, liaoxiangyu1, liux, limeng\_scu\}@stu.scu.edu.cn, chaoren@scu.edu.cn}
}

\begin{document}
\maketitle
\begin{abstract}
All-in-one image restoration aims to recover clear images from various degradation types and levels with a unified model. Nonetheless, the significant variations among degradation types present challenges for training a universal model, often resulting in task interference, where the gradient update directions of different tasks may diverge due to shared parameters. To address this issue, motivated by the routing strategy, we propose DFPIR, a novel all-in-one image restorer that introduces Degradation-aware Feature Perturbations(DFP) to adjust the feature space to align with the unified parameter space. In this paper, the feature perturbations primarily include channel-wise perturbations and attention-wise perturbations. Specifically, channel-wise perturbations are implemented by shuffling the channels in high-dimensional space guided by degradation types, while attention-wise perturbations are achieved through selective masking in the attention space. 
To achieve these goals, we propose a Degradation-Guided Perturbation Block (DGPB) to implement these two functions, positioned between the encoding and decoding stages of the encoder-decoder architecture.
Extensive experimental results demonstrate that DFPIR achieves state-of-the-art performance on several all-in-one image restoration tasks including image denoising, image dehazing, image deraining, motion deblurring, and low-light image enhancement. Our codes are available at \textcolor{cyan}{https://github.com/TxpHome/DFPIR}.

\end{abstract}    
\section{Introduction}
\label{sec:intro}

\begin{figure}[t]
	\setlength{\abovecaptionskip}{0.cm}
	\centering
	\includegraphics[width=1\linewidth]{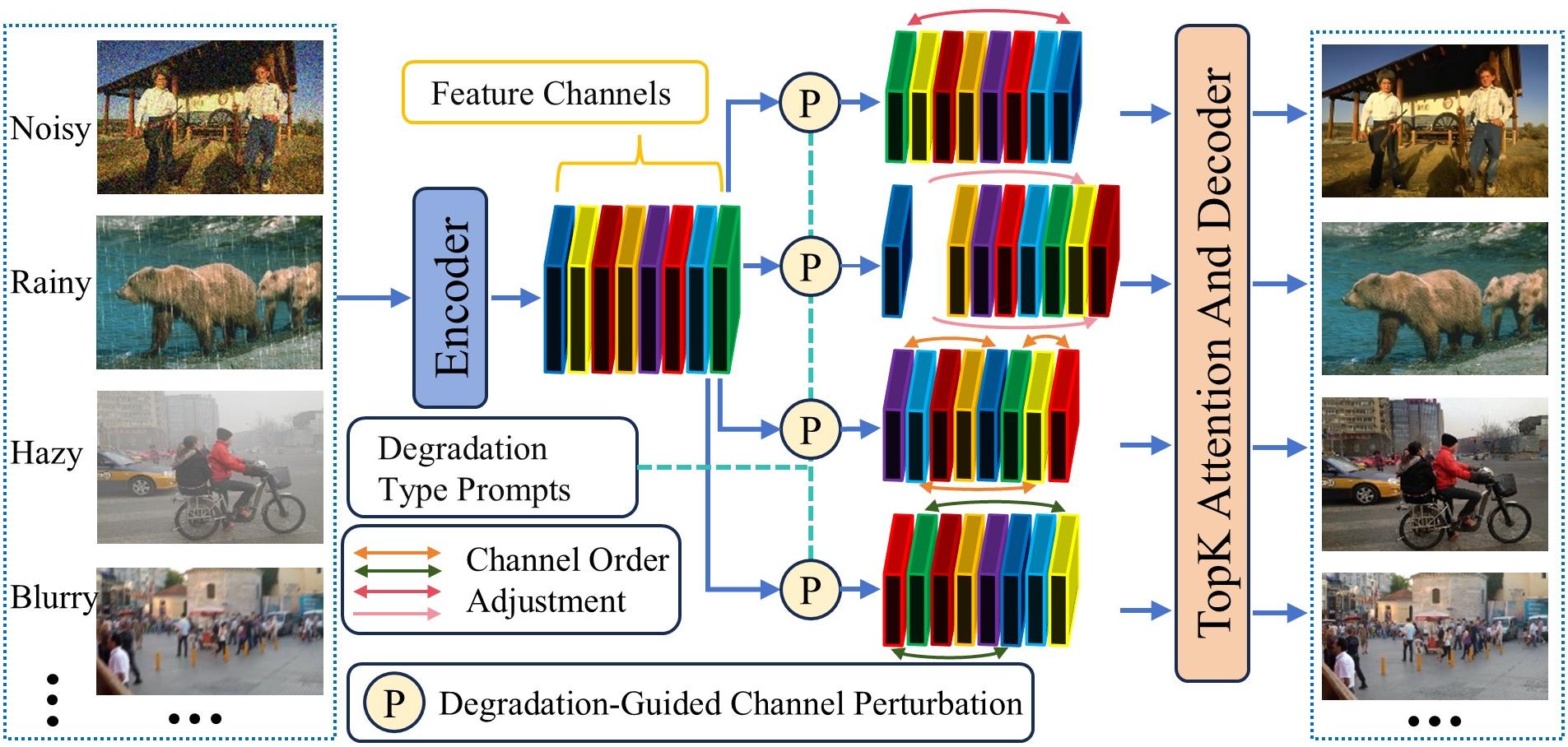} 
	\caption{This figure demonstrates the channel-wise perturbation method of DFPIR, where channel shuffle assigns unique channel orders for different degradation types.}
	\label{shuffle}
\end{figure}

Image restoration, a fundamental task in computer vision, has been widely studied, with a primary focus on addressing individual degradation types, such as noise, haze, rain, or blur. Recently, deep learning-based approaches have achieved remarkable progress in single-degradation restoration tasks, including denoising \cite{Denoise_DeamNet, Denoise_pan2023random, Denoise_lin2023unsupervised, Denoise_huang2021neighbor2neighbor, Denoise_zhussip2019training, Denoise_pan2022real}, deraining \cite{Derain_chen2021robust, Derain_li2019heavy, Derain_jiang2020multi, Derain_wang2020model, Derain_yang2020single}, dehazing \cite{Dehaze_chen2020pmhld, Dehaze_qin2020ffa, Dehaze_shao2020domain, Dehaze_das2020fast, Dehaze_wu2021contrastive}, and deblurring \cite{Deblur_cho2021rethinking, Deblur_kupyn2019deblurgan, Deblur_park2020multi, Deblur_zhang2020deblurring, Deblur_park2020multi}. While these single-degradation solutions perform well under specific conditions, they often struggle to generalize effectively to multiple degradation scenarios. Consequently, recent research has shifted toward multi-degradation restoration techniques, delivering state-of-the-art performance for known combinations of degradations \cite{IM_NAFNetchen2022simple, IM_mou2022deep, IM_zamir2022restormer, IM_zamir2021multi}. However, these approaches typically require separate networks for each degradation type, leading to large model sizes and high computational demands.

Recently, all-in-one approaches(also known as multi-degradation or multi-task image restoration) have gained prominence by addressing multiple image degradations within a unified model \cite{all-in-one_AirNet_li2022all, all-in-one_IDR_zhang2023ingredient, all-in-one_chen2022learning, potlapalli2023promptir,park2023all,yang2024all,wang2023smartassign}. While these methods have achieved state-of-the-art results, they often overlook the relationships and distinctive characteristics among different degradation types due to shared parameters. For example, MedIR \cite{yang2024all} demonstrates that the gradient update directions between different tasks are inconsistent or even opposite. According
to the characteristics of existing methods, current all-in-one
image restoration methods can be roughly divided into two
categories: (1) one solution is to modify the parameter space to fit the model for different degradations \cite{all-in-one_AirNet_li2022all,all-in-one_IDR_zhang2023ingredient,park2023all,wang2023smartassign}; (2) and the other solution is to modify the feature space to align with the shared parameter space \cite{potlapalli2023promptir,conde2024high-InstructIR,yang2024all}. Both approaches require incorporating degradation information into the network to mitigate the interference between different degradations. Although the approach (1) can effectively enhance network performance, it usually requires a large number of additional degradation parameters or a more complex network structure, increasing computational complexity. 

Compared to the approach (1), the method (2) is more prevalent in the “All-in-One” image restoration framework. The approach (2) typically involves introducing a degradation type prompt, which modulates features to align with the shared network parameters. Specifically, PromptIR \cite{potlapalli2023promptir} conducts multi-degradation processing by introducing additional implicit prompts. 
However, although this implicit cue in the feature domain utilizes the inherent features of the image, it neglects the influence of degradation types, making it challenging to reduce the mutual influence among multiple degradation types, which ultimately leads to suboptimal results.
MedIR \cite{yang2024all} introduces multiple experts into the feature space modulation to implement a task-adaptive routing strategy. Although this ``hard" routing strategy effectively reduces the influence between tasks, it may overlook the inherent features of the image among multiple degradations. InstructIR \cite{conde2024high-InstructIR} introduces text commands for multi-task restoration, showcasing the potential of using text prompts to guide image restoration. However, InstructIR  \cite{conde2024high-InstructIR} modulates features through channel attention using text prompts, which may make it challenging to mitigate the mutual influence among different degradations.

\begin{figure}[t]
	\setlength{\abovecaptionskip}{0.cm}
	\centering
	\includegraphics[width=1\linewidth]{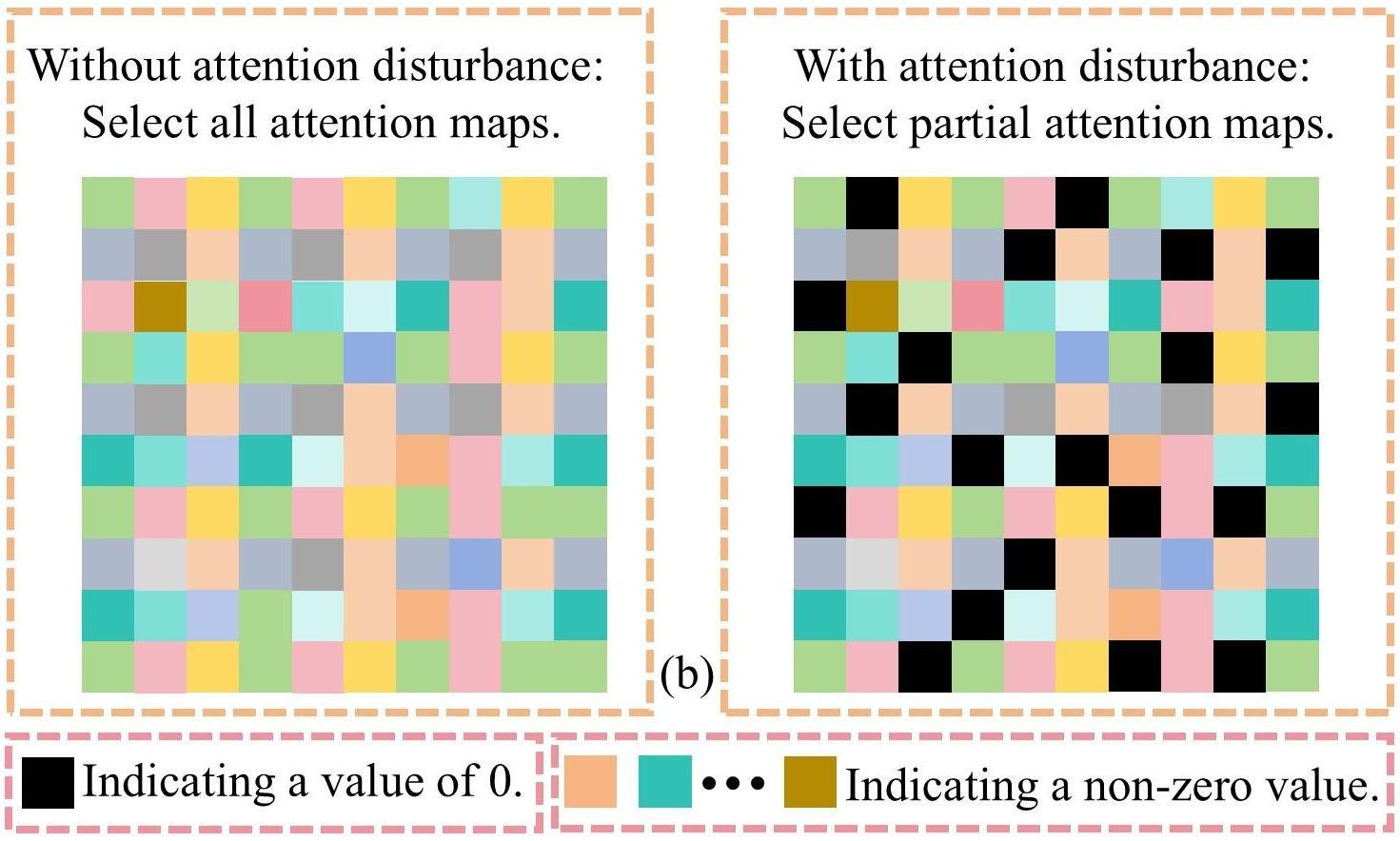} 
	\caption{This figure showcases our attention-wise perturbation, where DFPIR applies attention selection to perturb image features, discarding a portion of attention for each degradation type.}
	\label{topkmask}
\end{figure}

To overcome these challenges, this paper analyzes the design philosophy and rationale behind the success of prompts and presents a degradation-aware feature perturbation approach to adjust the feature space to align with the unified parameter space for all-in-one image restoration. Our method modulates features through perturbations which include channel-wise and attention-wise perturbations to align with shared network parameters or structures, guided by degradation type prompts. Specifically, channel-wise perturbations involve shuffling the channels in high-dimensional space rather than using the conventional channel attention mechanism (see Fig.\ref{shuffle}). 
This shuffling approach preserves the inherent features of the image while reducing the mutual influence of degradation features.
While attention-wise perturbations involve selectively masking the features after channel shuffling to achieve channel-adapted attention perturbation (see Fig.\ref{topkmask}). 
This approach not only preserves the inherent feature information of the image but also adaptively reduces the mutual influence between different degradation types, achieving a good balance between influence and inherent features. 
The main contributions of our method are as follows:
\begin{itemize}
	\item We propose a novel degradation-aware feature perturbation all-in-one restoration framework, DFPIR, which adaptively adjusts the feature space in high-dimensional space to align with the unified parameter space, guided by degradation type prompts.
		
	\item We design a Degradation-Guided Perturbation Block (DGPB), consisting of a Degradation-Guided Channel Perturbation Module (DGCPM) and a Channel-Adapted Attention Perturbation Module (CAAPM), to apply perturbation modulation to features along the channel and attention dimensions, aligning with the encoder-decoder architecture with unified parameters.	
	
	\item Extensive experiments demonstrate that our network achieves state-of-the-art performance in all-in-one image restoration. Especially for all-in-one restoration, 0.45dB PSNR improvement is obtained in comparison with InstructIR \cite{conde2024high-InstructIR}.
\end{itemize}

\section{Related Works}
\label{relate works}

\paragraph{Single Task Image Restoration.}
Single task image restoration seeks to recover a high-quality image from its low-quality version. Due to the ill-posed nature of the task, early methods primarily focused on designing effective handcrafted priors to constrain the solution space. In contrast, 
recent deep learning-based approaches have significantly advanced the performance of various image restoration tasks, such as denoising \cite{Denoise_DeamNet, Denoise_ScaoedNet, Denoise_pan2023random, Denoise_lin2023unsupervised, Denoise_huang2022casapunet, Denoise_pan2022real}, deraining \cite{Derain_chen2021robust, Derain_li2018recurrent, Derain_wang2019erl, Derain_li2019heavy, lin2023unlocking}, dehazing \cite{Dehaze_chen2020pmhld, Dehaze_liu2019griddehazenet, Dehaze_qin2020ffa, Dehaze_shao2020domain, Dehaze_das2020fast, Dehaze_wu2021contrastive}, deblurring \cite{chen2024unsupervised,Deblur_cho2021rethinking, Deblur_kupyn2018deblurgan, Deblur_kupyn2019deblurgan, Deblur_park2020multi, Deblur_zhang2020deblurring, Deblur_park2020multi}, and low-light enhancement \cite{Lowlight_guo2020zero, Lowlight_ma2022toward, Lowlight_wu2022uretinex}, by learning generalizable priors from large-scale datasets.
We focus on general purpose restoration models \cite{liang2021swinir,IM_NAFNetchen2022simple, IM_zamir2022restormer}, as these architectures can be independently trained for a variety of tasks. 
NAFNet \cite{IM_NAFNetchen2022simple} simplifies the network structure with lightweight channel attention and gated mechanisms, offering an alternative to non-linear activations. Meanwhile, Restormer \cite{IM_zamir2022restormer} leverages transformer architectures to enhance low-level restoration tasks while minimizing computational overhead.
In this study, we adopt Restormer \cite{IM_zamir2022restormer} as the backbone for our DFPIR model due to its efficient design and high performance across multiple restoration tasks. 
However, these models are primarily designed for single-degradation scenarios, limiting their effectiveness when applied directly to all-in-one restoration tasks.

\paragraph{All-In-One Image Restoration.} 
Multi-task image restoration aims to address multiple tasks using the same network design \cite{all-in-one_AirNet_li2022all,all-in-one_IDR_zhang2023ingredient,park2023all,wang2023smartassign,potlapalli2023promptir,conde2024high-InstructIR,yang2024all}. Compared to single-task image restoration, the key challenge in multi-task image restoration is how to reduce the mutual influence of different degradation features while preserving the inherent features of the image.
One approach is to adjust the network's parameter space to accommodate different degradation types. 
Several works have explored these strategies to handle diverse degradations. Li et al. \cite{all-in-one_li2020all} introduce a single-encoder, multi-decoder framework targeting weather-based degradations using the Rain-Haze-Snow dataset. Chen et al. \cite{all-in-one_chen2022learning} propose a two-stage knowledge transfer mechanism, employing a multi-teacher, single-student approach to handle various degradation types. 
Li et al. \cite{all-in-one_AirNet_li2022all} propose an all-in-one framework capable of restoring multiple degraded images without requiring prior knowledge of degradation types or levels. Zhang et al. \cite{all-in-one_IDR_zhang2023ingredient} propose an ingredient-oriented strategy that supports up to five restoration tasks within a single model, significantly enhancing scalability. Similarly, Zhang et al. \cite{zhang2023all} introduce a representation learning network guided by degradation classification, using its strong classification capabilities to effectively steer the restoration process.  
Another approach is to introduce image or degradation prompts to modulate the features, adapting to a unified parameter space \cite{potlapalli2023promptir,conde2024high-InstructIR,yang2024all}. 
PromptIR \cite{potlapalli2023promptir} encodes degradation-specific information through prompts, using them to dynamically guide the restoration network. In contrast, InstructIR \cite{conde2024high-InstructIR} allows for model-driven image editing based on instructions that specify the desired actions, rather than relying on text labels, captions, or descriptions of input and output images. 

\section{Proposed Method}
\label{method}
\begin{figure*}[htbp]
	\setlength{\abovecaptionskip}{0.cm}
	\centering
	\includegraphics[width=1\linewidth]{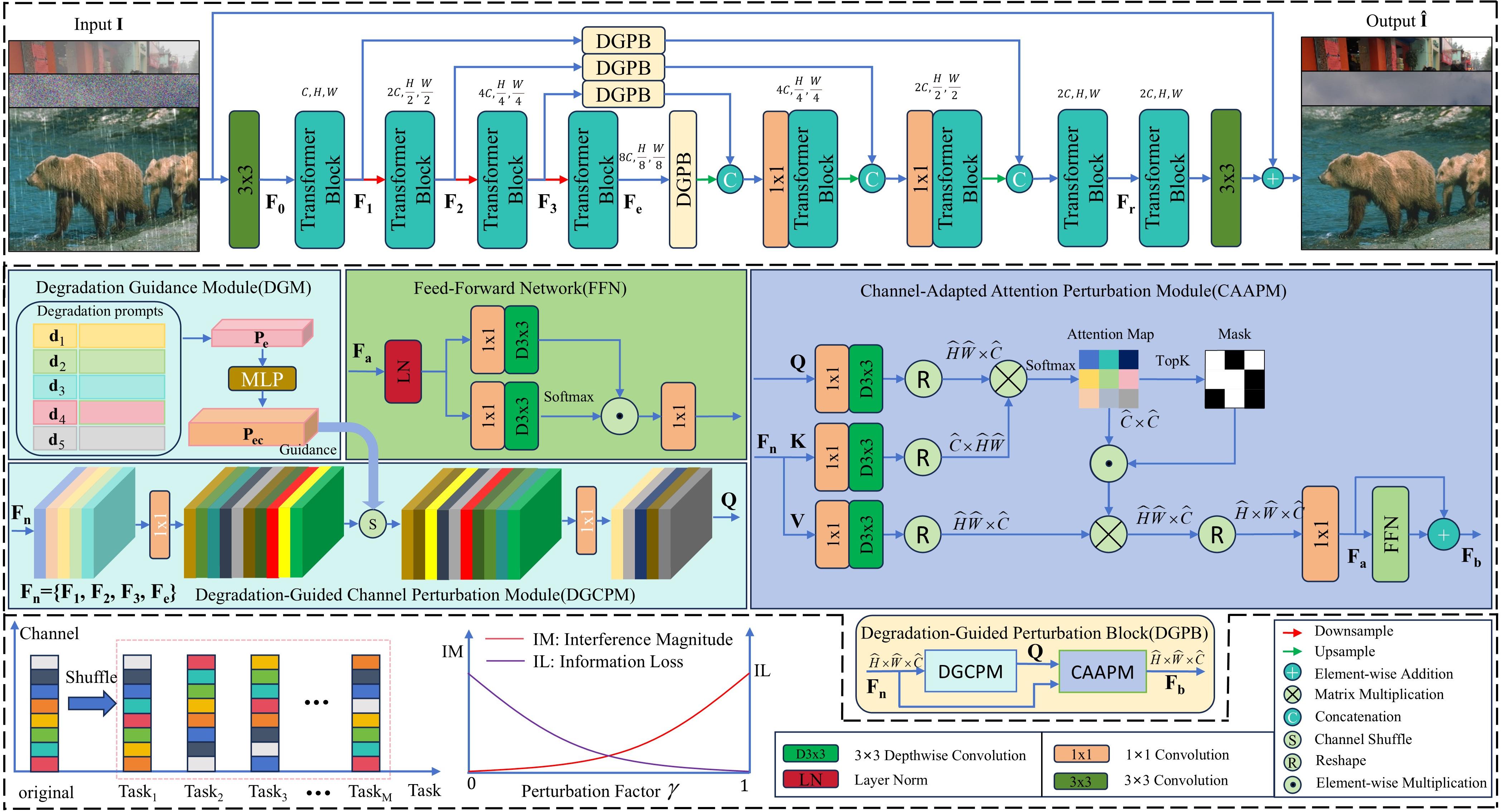}
	\caption{Overview of the DFPIR framework. We employ Restormer \cite{IM_zamir2022restormer}, an encoder-decoder network with transformer blocks in the encoding and decoding stages, as our backbone. The primary component of the framework, the Degradation-Guided Perturbation Block (DGPB), consists of two submodules, i.e., Degradation-Guided Channel Perturbation Module (DGCPM) and Channel-Adapted Attention Perturbation Module (CAAPM). The DGCPM module introduces conventional dimensional perturbations to image features in the form of channel shuffling, guided by degradation-type prompts. The CAAPM module applies attention perturbation to the channel-shuffled features through a top-K masking strategy.}
	\label{overall framework}
\end{figure*}

In ``All-In-One" image restoration, the objective is to develop a single model $\textbf{M}$ capable of restoring a clean image $\widehat{\textbf{I}}$ from a degraded input image $\textbf{I}$ affected by a degradation $\textbf{D}$.
Within this framework, under the same network parameters and architecture, different types of degradation mutually influence each other. For example, degraded images contain both inherent image features and degradation features. While inherent image features help the network learn latent parameters, thereby enhancing restoration performance, degradation features may negatively impact each other, reducing restoration effectiveness.
In order to fully utilize the inherent features of the image while reducing the influence of different degradation features, motivated by the ``hard" routing strategy \cite{yang2024all} and PromptIR \cite{potlapalli2023promptir}, we propose a degradation-aware feature perturbation modulation network, DFPIR, that adjusts the feature space through perturbation modulation to align with the shared-parameter network structure, thereby enhancing multi-degradation image restoration performance. The perturbation is divided into two parts: channel-wise perturbation and attention-wise perturbation. Channel-wise perturbation is implemented through channel shuffle, while attention-wise perturbation is achieved by selecting a portion of the attention maps.
The details of these two perturbation mechanisms are provided in Section \ref{analasis_shuffle}.
The pipeline of the proposed DFPIR framework is illustrated in Fig.\ref{overall framework}. In the following sections, we describe the overall structure of DFPIR in detail.

\begin{figure}[t]
	\setlength{\abovecaptionskip}{0.cm}
	\centering
	\includegraphics[width=1\linewidth]{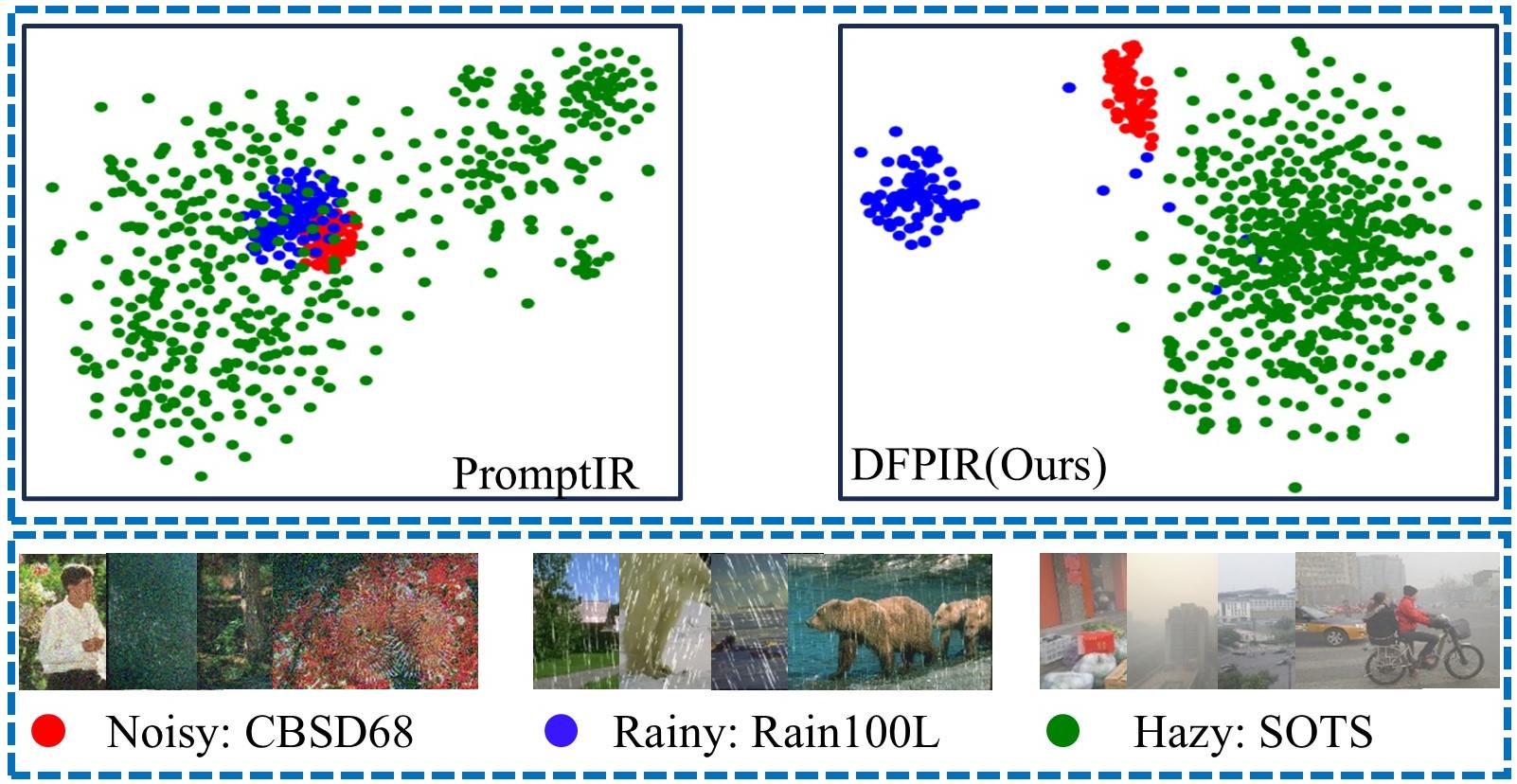} 
	\caption{The figure provides t-SNE plots of the intermediate features from DFPIR (our method) and PromptIR on the test datasets (CBSD68, Rain100L, and SOTS) under the three-task setting. In our model, the features for each task exhibit tighter clustering, highlighting the effectiveness of our degradation-aware feature perturbation strategy in enhancing restoration performance.}
	\label{tsne-3d}
\end{figure}

\subsection{Overall Pipeline}
Given a degraded input image ${\textbf{I}} \in {R^{H \times W \times 3}}$, the DFPIR initially extracts shallow features ${\textbf{F}_\textbf{\small 0}} \in {R^{H \times W \times C}}$ using a $3 \times 3$ convolution layer; where $H \times W$ is the spatial size and $C$ denotes the channels. Next, these features ${\textbf{F}_\textbf{\small 0}}$ undergo a 4-level encoder-decoder network, transforming into deep features ${\textbf{F}_\textbf{\small r}} \in {R^{H \times W \times 2C}}$. Each level of the encoder-decoder employs multiple Transformer blocks \cite{IM_zamir2022restormer}, where the number of blocks gradually increases from the top level to the bottom level, facilitating a computationally efficient design. Beginning with the high-resolution input, the encoder aims to gradually decrease spatial resolution while increasing channel capacity, resulting in a low-resolution latent representation ${\textbf{F}_\textbf{\small e}} \in {R^{\frac{H}{8} \times \frac{W}{8} \times 8C}}$. Given the low-resolution latent features ${\textbf{F}_\textbf{\small e}}$, the objective of the decoder is to progressively restore the high-resolution clean output. To assist the recovery process, the encoder features are concatenated with the decoder features via skip connections \cite{ronneberger2015u}.
We insert our Degradation-Guided Perturbation Block (DGPB) between the encoder and decoder, specifically within the skip-connection stage, to perturb the encoded feature space, which is guided by degradation type prompts, and align it with the shared-parameter decoder. 
To obtain degradation type prompts, we utilize a pre-trained CLIP \cite{radford2021learning} model to encode the textual degradation type descriptions. In the following sections, we describe the proposed DGPB and its core building modules in detail.

\subsection{Degradation-Guided Perturbation Block (DGPB)}
In all-in-one setting, shared parameters are challenging to handle all-in-one image restoration effectively. We specifically design a Degradation-Guided Perturbation Block (DGPB) to apply perturbations to the encoded features, better aligning them with the shared decoder in multi-task scenarios, guided by the degradation type prompts (as shown in Fig.\ref{overall framework}). Given as inputs both the image feature $ {\textbf{F}_\textbf{\small n}} \in {R^{ \widehat H \times \widehat W \times \widehat C }} $ and degradation type prompts $ {\textbf{P}_\textbf{\small e}} \in {T^{\widehat L \times 1}} $, the overall process of DGPB is defined as: 
\begin{equation}
	{{\textbf{F}}_{\textbf{\small b}}} = {\rm{CAAPM}}({\rm{DGCPM}}({{\textbf{F}}_{\textbf{\small n}}},{\rm{DGM}}({{\textbf{P}}_{\textbf{\small e}}})),{{\textbf{F}}_{\textbf{\small n}}}) 
	\label{eq1}
\end{equation}
where $ {\textbf{F}_\textbf{\small n}} $ represents the encoder output. The DGPB comprises two core components: the Degradation-Guided Channel Perturbation Module (DGCPM) and the Channel-Adapted Attention Perturbation Module (CAAPM).

\begin{table*}[htbp]
	\caption{Comparison to state-of-the-art on three tasks. PSNR (dB, ↑) and SSIM (↑) metrics	are reported on the full RGB images. On average PSNR, our DFPIR provides a significant gain of 0.45 dB over the previous all-in-one method InstructIR \cite{conde2024high-InstructIR}.}
	\label{all-in-one-3T}
	\centering
		\setlength{\tabcolsep}{8.5pt}
	\begin{tabular}{c|ccccc|c} 
		\hline  
		\multirow{2}{*}{Method}  & Dehazing  & Deraining & \multicolumn{3}{c|}{Denoising on CBSD68 dataset}  & \multirow{2}{*}{Average}                \\
		& on SOTS  & on Rain100L & $\sigma = 15$& $\sigma = 25$ & $\sigma = 50$ &         \\
		\hline 
		DL \cite{fan2019general}   &26.92 / 0.391&32.62 / 0.931&33.05 / 0.914&30.41 / 0.861&26.90 / 0.740&29.98 / 0.875 \\  
		FDGAN \cite{dong2020fd}     &24.71 / 0.924&29.89 / 0.933&30.25 / 0.910&28.81 / 0.868&26.43 / 0.776&28.02 / 0.883 \\
		MPRNet \cite{IM_zamir2021multi}  &25.28 / 0.954&33.57 / 0.954&33.54 / 0.927&30.89 / 0.880&27.56 / 0.779&30.17 / 0.899 \\ 
		AirNet \cite{all-in-one_AirNet_li2022all}  &27.94 / 0.962&34.90 / 0.967&33.92 / 0.933&31.26 / 0.888&28.00 / 0.797&31.20 / 0.910 \\ 
		Restormer \cite{IM_zamir2022restormer}  &30.43 / \textcolor{blue}{0.975}&36.55 / 0.975&33.84 / 0.931&31.18 / 0.885&27.90 / 0.790&31.98 / 0.911 \\ 
		PromptIR \cite{potlapalli2023promptir}  &\textcolor{blue}{30.58} / 0.974&36.37 / 0.972&33.98 / 0.933&31.31 / 0.888&28.06 / 0.799&32.06 / 0.913 \\ 
		InstructIR \cite{conde2024high-InstructIR}  &30.22 / 0.959 & \textcolor{blue}{37.98 / 0.978} &\textcolor{red}{34.15} / \textcolor{blue}{0.933} &\textcolor{red}{31.52} / \textcolor{blue}{0.890} &\textcolor{red}{28.30} / \textcolor{blue}{0.804} &\textcolor{blue}{32.43 / 0.913} \\
		\hline				
		DFPIR(Ours) 
		&\textcolor{red}{31.87 / 0.980}&\textcolor{red}{38.65 / 0.982}& \textcolor{blue}{34.14} / \textcolor{red}{0.935} & \textcolor{blue}{31.47} / \textcolor{red}{0.893}  & \textcolor{blue}{28.25} / \textcolor{red}{0.806} &  \textcolor{red}{32.88 / 0.919} \\		
		\hline		
	\end{tabular}
\end{table*}

\subsubsection{Degradation-Guided Channel Perturbation Module (DGCPM)}
In DGCPM, the objective is to add perturbations to the feature channels through channel shuffling, guided by degradation type prompts. However, directly shuffling the channels on the feature $ {\textbf{F}_\textbf{\small n}} $ may make it difficult to converge due to excessive perturbation and may also affect the quality of the reconstruction. To address this issue, we initially expand the number of channels in the image feature $ {\textbf{F}_\textbf{n}}$ by 2x to introduce channel perturbations in a higher-dimensional channel space, guided by the Degradation Guidance Module(DGM). After channel expansion, the size of feature $ {\textbf{F}_\textbf{\small n}}$ becomes  $ {\textbf{F}_\textbf{\small 2n}} \in {R^{\widehat H \times \widehat W \times 2 \widehat C}} $. 
In DGM, the degradation type prompts ($ {\textbf{P}_\textbf{\small e}} $) are used to adaptively apply channel shuffle.
To maintain the dimensionality of $ {\textbf{P}_\textbf{\small e}}$ consistent with the channel dimensionality of $ {\textbf{F}_\textbf{\small 2n}}$, we employ a Multi-Layer Perceptron (MLP) which consisting of two linear layers to achieve dimension matching. 
Thus, MLP  transforms the input feature $ {\textbf{P}_\textbf{\small e}}$ into $ {\textbf{P}_\textbf{\small ec}} \in {T^{2 \widehat C \times 1}} $. \textbf {For degradation type prompts, see the supplementary document.}
In channel shuffle stage, we get the index value corresponding to the top-K ($ \rm{K}= 2 \widehat C$) value of $ {\textbf{P}_\textbf{\small ec}} $, and use these index values to reorder the channels. Finally, halve the number of channels after the shuffle to keep the channel count consistent before and after the transformation. Overall, the DGCPM process is summarized as:
\begin{equation}
{\textbf{Q}} = Convh({\textbf{C}}{{\textbf{S}}_{topK}}(Convk({{\textbf{F}}_{\textbf{n}}})|{\rm{DGM}}({{\textbf{P}}_{\textbf{e}}})))
\label{eq2}
\end{equation}	
where the operation \scalebox{0.9}{${\textbf{C}}{{\textbf{S}}{topK}}\left( {\bullet|{\rm{DGM}}({{\textbf{P}}{\textbf{e}}})} \right)$} denotes top-K channel shuffling guided by $ {\rm{DGM}}({{\textbf{P}}_{\textbf{e}}})$, while \scalebox{0.9}{$ Convk\left( {\bullet} \right) $} and \scalebox{0.9}{$ Convh\left( {\bullet} \right) $} represent the operations of doubling and halving the feature channels, respectively. After DGCPM, we obtain feature $ {\textbf{Q}} \in {R^{\widehat H \times \widehat W \times \widehat C}} $.

\subsubsection{Channel-Adapted Attention Perturbation Module(CAAPM)}
Although the shuffled features can adapt well to specific degradations, directly using them for reconstruction may not yield optimal results. This is because the shuffled features carry degradation-type information but lack interaction with the original feature information. To address this, we designed the Channel-Adapted Attention Perturbation Module (CAAPM). CAAPM has two main functions: facilitating information interaction between the shuffled and original features, and adding perturbations in the attention dimension. Motivated by  Restormer \cite{IM_zamir2022restormer}, we design a cross-attention mechanism in the channel dimension to aggregate the shuffled and original features. To add perturbations to the attention map, we introduce a mask matrix ${\textbf{M}} \in {R^{\widehat C \times \widehat C}}$ to select part of the attention map from each row using a top-K approach with a perturbation factor parameter $ \gamma $. The process of obtaining the perturbed attention map ${\textbf{PAM}} \in {R^{\widehat H \widehat W \times \widehat C}}$ can be described as:
\begin{equation}\label{eq_qkv} 
{\textbf{PAM}}({\textbf{Q,K,V}}) = {\rm{softmax}}(\textbf{M} \odot ({{{{\textbf{Q}}_{\rm{*}}}{\textbf{K}}_*^T} \mathord{\left/
		{\vphantom {{{{\rm{Q}}_{\rm{*}}}{\rm{K}}_*^T} {\sqrt {{d_k}} }}} \right.
		\kern-\nulldelimiterspace} {\sqrt {{d_k}} }})){{\textbf{V}}_*}
\end{equation}
where the query $ {{\textbf{Q}}_*} $ is derived from shuffled feature $ \textbf{Q} $, and the key $ {{\textbf{K}}_*} $ and value  $ {{\textbf{V}}_*} $ are derived from original features $ {\textbf{F}_\textbf{\small n}} $. $ {\textbf{M}} $ represents the selection mask matrix. $  \odot$ represents the element-wise multiplication. Then, we use a 1×1 convolution to obtain the feature $ {{\textbf{F}}_{\textbf{a}}} $ with attention perturbations. Finally, we obtain the final output through an FFN network. This can be expressed as:	
\begin{equation}
{{\textbf{F}}_{\textbf{b}}} = {{\textbf{F}}_{\rm{a}}} + \rm{FFN}({{\textbf{F}}_{\textbf{a}}})	
\end{equation}	

The selection of the perturbation factor $ \gamma $ needs to comprehensively consider the extent of interference magnitude between tasks and information loss. Fig.\ref{overall framework} illustrates the relationship between $ \gamma $, interference size, and information loss. 
In our paper, the parameter is $ \gamma $ fixed at 0.9, and the ablation study presents the experimental results for different values of $ \gamma $.

\begin{table*}[t]
	\caption{ Comparison to state-of-the-art on five tasks. PSNR (dB, ↑) and SSIM (↑) metrics	are reported on the full RGB images with (*) denoting general image restorers, others are specialized all-in-one approaches. Denoising results are reported for the noise level $ \sigma = 25 $. }
	\label{all-in-one-5T}
	\centering
	\setlength{\tabcolsep}{6.8pt} 
	\begin{tabular}{c|ccccc|c}
		\hline  
		\multirow{2}{*}{Method} & Dehazing & Deraining & Denoising& Deblurring & Low-light Enh.& \multirow{2}{*}{Average} \\
		& on SOTS  & on Rain100L  & on CBSD68  & on Gopro  & on LOL  &  \\
		\hline 
		DGUNet* \cite{IM_mou2022deep}  &24.78 / 0.940&36.62 / 0.971&31.10 / 0.883&27.25 / 0.837&21.87 / 0.823&28.32 / 0.891 \\ 
		SwinIR* \cite{liang2021swinir}   &21.50 / 0.891&30.78 / 0.923&30.59 / 0.868&24.52 / 0.773&17.81 / 0.723&25.04 / 0.835 \\ 
		Restormer* \cite{IM_zamir2022restormer} &24.09 / 0.927&34.81 / 0.962&31.49 / 0.884&27.22 / 0.829&20.41 / 0.806&27.60 / 0.881 \\ 
		NAFNet* \cite{IM_NAFNetchen2022simple}   &25.23 / 0.939&35.56 / 0.967&31.02 / 0.883&26.53 / 0.808&20.49 / 0.809&27.76 / 0.881 \\ 
		\hline 	
		DL \cite{fan2019general} &20.54 / 0.826&21.96 / 0.762&23.09 / 0.745&19.86 / 0.672&19.83 / 0.712&21.05 / 0.743 \\ 
		Transweather \cite{valanarasu2022transweather}&21.32 / 0.885&29.43 / 0.905&29.00 / 0.841&25.12 / 0.757&21.21 / 0.792&25.22 / 0.836 \\ 
		TAPE \cite{liu2022tape}  &22.16 / 0.861&29.67 / 0.904&30.18 / 0.855&24.47 / 0.763&18.97 / 0.621&25.09 / 0.801 \\
		AirNet \cite{all-in-one_AirNet_li2022all}    &21.04 / 0.884&32.98 / 0.951&30.91 / 0.882&24.35 / 0.781&18.18 / 0.735&25.49 / 0.846 \\ 
		IDR \cite{all-in-one_IDR_zhang2023ingredient}      &25.24 / 0.943&35.63 / 0.965&\textcolor{red}{31.60} / \textcolor{blue}{0.887} &27.87 / 0.846&21.34 / 0.826&28.34 / 0.893 \\ 
		InstructIR \cite{conde2024high-InstructIR}  &\textcolor{blue}{27.10 / 0.956}& \textcolor{blue}{36.84 / 0.973} & \textcolor{blue}{31.40 / 0.887}  &\textcolor{red}{29.40 / 0.886}& \textcolor{blue}{23.00 / 0.836}& \textcolor{blue}{29.55 / 0.907} \\
		
		\hline 				
		DFPIR(Ours) 
		&\textcolor{red}{31.64 / 0.979}&\textcolor{red}{37.62 / 0.978}&31.29 / \textcolor{red}{0.889} &\textcolor{blue}{28.82 / 0.873} &\textcolor{red}{23.82 / 0.843} &\textcolor{red}{30.64 / 0.913} \\ 		
		\hline		
	\end{tabular}				
\end{table*}

\subsection{Analysis of Our Feature Perturbation Strategy}
The task-aware feature perturbation strategy we propose involves both channel and attention dimensions. By perturbing these two dimensions, it not only preserves the inherent features of the image but also reduces the mutual influence of degradation features.
Fig.\ref{tsne-3d} illustrates that our model excels in learning discriminative degradation contexts. 
\label{analasis_shuffle}

\paragraph{Channel-wise Perturbation Strategy.} Directly performing channel shuffle in a low-dimensional space causes excessive perturbation, making network training difficult to converge. 
Therefore, we propose a degradation-aware adaptive channel shuffle strategy (as shown in Fig.\ref{shuffle}), which adaptively reorders the feature channels for each task in a high-dimensional feature space.
Given a image feature which contains $ N $ channels $ {{\textbf{F}}_{{\textbf{\small nc}}}} = \left\{ {{C_1},{C_2}, \cdots ,{C_r},\cdots ,{C_N}} \right\} {\rm{(r}} \le {N)} $ and $ \rm{M} $ tasks, our channel shuffle strategy adopts different channel orders for each degradation restoration task. 
For any given task $ {{\rm{T}}_{\rm{m}}}{\rm{(m}} \le {\rm{M)}} $, the channel index vector after channel shuffle is defined as ${{\bf{S}}^N} = \{ S_1^N,S_2^N, \cdots ,S_N^N\} $. Finally, the feature obtained after channel shuffle is
${{\textbf{F}}_{\textbf{m}}} = \{ {C_{S_1^N}},{C_{S_2^N}}, \cdots ,{C_{S_N^N}}\} $. 
The adjustment of this channel order is adaptively random under the guidance of degradation prompts.
This random adaptive shuffle strategy reduces interference between feature channels while preserving the inherent characteristics of the image (as channel shuffling does not result in channel feature loss), thereby improving reconstruction quality.

\begin{figure*}[htbp]
	\setlength{\abovecaptionskip}{0.cm}
	\centering
	\includegraphics[width=1\linewidth]{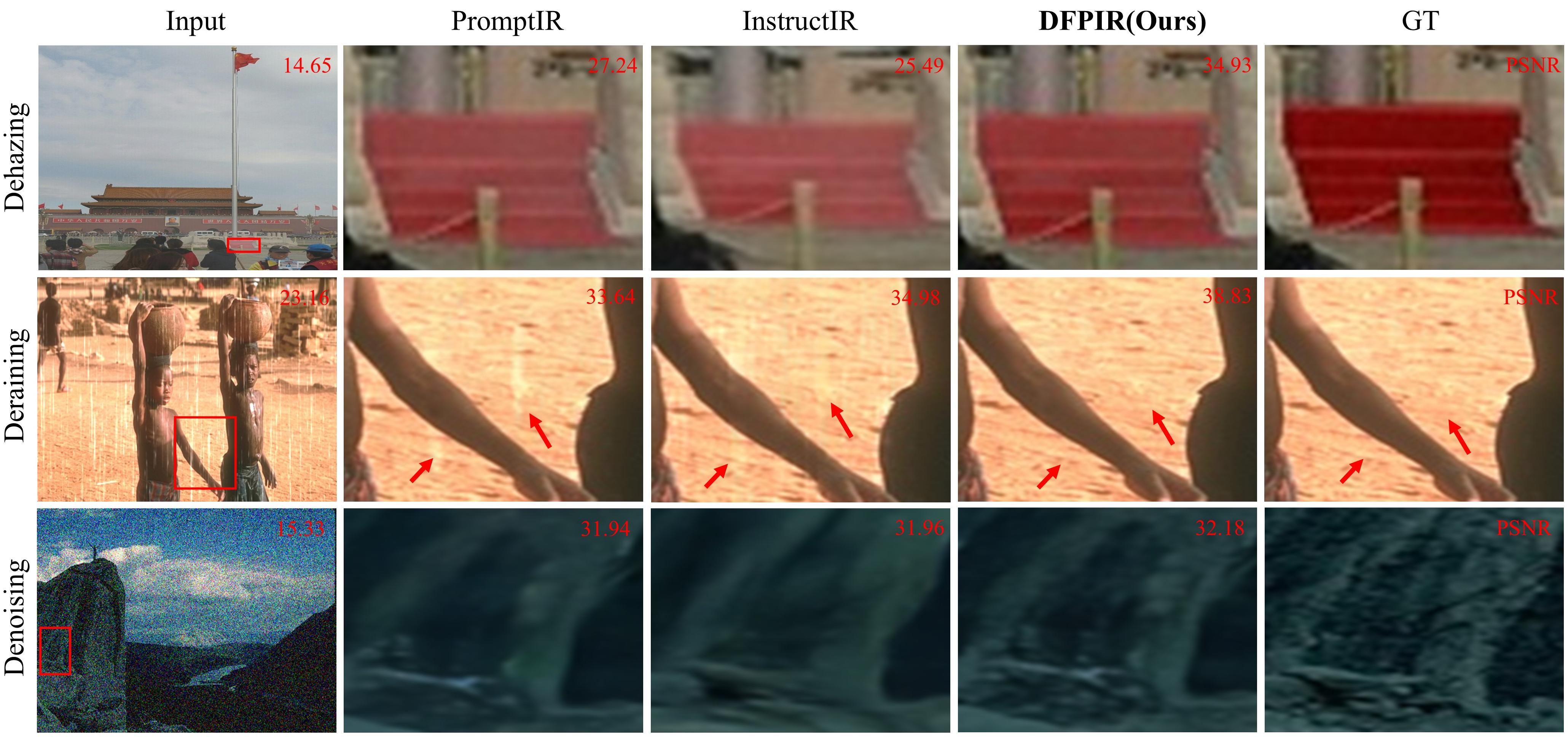}  
	\caption{Visual comparison of DFPIR with state-of-the-art methods on challenging cases for the All-in-One setting considering three degradations. Zoom in for better view.}
	\label{visual result-3t}
\end{figure*}

\paragraph{Attention-wise Perturbation Strategy.} 
The main purpose of channel shuffle is to minimize the influence of degradation features while preserving the inherent characteristics of the image. However, this reduction is not thorough enough, meaning its effectiveness is limited. To address this, we directly discard a portion of the attention in the attention maps to further mitigate the impact of multiple degradation features (as shown in Fig.\ref{topkmask}). Specifically, we compute the transposed cross attention map $ {\textbf{A}_{tt}} \in {R^{\widehat C \times  \widehat C}} $. 
We select a portion of the attention map values from each column of 
$ {\textbf{A}_{tt}} $ using the top-K method to generate a disturbance mask matrix $ {\textbf{M}} \in {R^{\widehat C \times  \widehat C}} $, where the values that are not selected are set to 0, and the remaining values are set to 1. Then, the matrix $ {\textbf{M}} $ is multiplied element-wise with $ {\textbf{A}_{tt}} $ to obtain a new attention map 
$ {\textbf{A}_{ttn}} $, achieving perturbation in the attention dimension. In other words, $ {{\bf{A}}_{ttn}} = {{\bf{A}}_{tt}} \odot {\bf{M}} $, where $ {\textbf{A}_{tt}} $ is ${{{{\bf{Q}}_{\rm{*}}}{\bf{K}}_*^T} \mathord{\left/
		{\vphantom {{{{\bf{Q}}_{\rm{*}}}{\bf{K}}_*^T} {\sqrt {{d_k}} }}} \right.
		\kern-\nulldelimiterspace} {\sqrt {{d_k}} }}$ in equation \ref{eq_qkv}.

\section{Experiments}
\label{experiments}

In this section, we follow the protocols of prior state-of-the-art works \cite{potlapalli2023promptir, all-in-one_AirNet_li2022all} to conduct experiments under two settings: (a) all-in-one and (b) single-task.
For the all-in-one setting, a unified model is trained to handle multiple degradation types, with experiments conducted across three and five distinct degradations. In contrast, the single-task setting involves training separate models, each specialized for a specific restoration task.
The image quality metrics—PSNR and SSIM \cite{huynh2008scope}—for the top-performing methods are highlighted in \textcolor{red}{red} and the second-best results are highlighted in \textcolor{blue}{blue} in the result tables. 
\textbf{Single-task results and more visual experiments are in the supplementary material.}

\subsection{Experimental Settings}
\textbf{Datasets.} In line with previous work \cite{potlapalli2023promptir, all-in-one_AirNet_li2022all}, we prepare datasets tailored for various restoration tasks. For single-task image denoising, we merge images from the BSD400 \cite{arbelaez2010contour} and WED \cite{ma2016waterloo} datasets to train the model. 
Testing is conducted on the CBSD68 \cite{martin2001database} and Urban100 \cite{huang2015single} datasets. For image dehazing, we use the SOTS \cite{li2018benchmarking} dataset, while Rain100L \cite{yang2020learning} is employed for image deraining. Deblurring and low-light enhancement tasks utilize the GoPro \cite{nah2017deep} and LOL-v1 \cite{nah2017deep} datasets, respectively.
In the all-in-one setting, a unified model is trained on the combined training datasets mentioned above and tested directly across multiple restoration tasks. 
\textbf{Additional dataset details can be found in the supplementary materials.}

\textbf{Implementation Details.} Our DFPIR provides an end-to-end trainable solution, removing the necessity for pretraining any individual components. Following the configuration of PromptIR \cite{potlapalli2023promptir}, our DFPIR architecture features a 4-level encoder-decoder structure, comprising varying numbers of Transformer blocks at each level, specifically [4, 6, 6, 8] from level-1 to level-4. We integrate our Degradation-Guided Perturbation Block (DGPB) between the encoder and decoder, with a total of four DGPBs distributed throughout the network.
We conduct our experiments using PyTorch on a single NVIDIA GeForce RTX 3090 GPU. For training, we run for 80 epochs with an initial learning rate of  $ 1{e^{ - 4}} $, followed by fine-tuning for 5 epochs at a learning rate of $ 1{e^{ - 5}} $. Initially, we set the patch size to $ 128^2 $ with the batch size of 5, and for fine-tuning, the patch size is adjusted to $ 192^2 $ with a batch size of 3.
The network is optimized using the L1 loss function in conjunction with the Adam optimizer (with parameters $ {\beta _1} = 0.9 $ and $ {\beta _2} = 0.999 $). We train on cropped patches and augment the dataset with random horizontal and vertical flips.

\subsection{Results Comparisons on Three Tasks}
We assess the performance of our all-in-one DFPIR across three distinct restoration tasks: dehazing, deraining, and denoising. Our DFPIR is compared with various general image restoration methods, including Restormer \cite{IM_zamir2022restormer}, FDGAN \cite{dong2020fd}, and MPRNet \cite{IM_zamir2021multi}, as well as specialized all-in-one approaches such as DL \cite{fan2019general}, AirNet \cite{all-in-one_AirNet_li2022all}, PromptIR \cite{potlapalli2023promptir}, and InstructIR \cite{conde2024high-InstructIR}.
As shown in Tab. \ref{all-in-one-3T}, the proposed DFPIR consistently outperforms the other competing methods. On average across the different restoration tasks, our algorithm achieves a performance improvement of \textcolor{red}{0.45} dB over the previous best method, InstructIR \cite{conde2024high-InstructIR}, and \textcolor{red}{0.82} dB over the second-best approach, PromptIR \cite{potlapalli2023promptir}. Specifically, DFPIR demonstrates a notable increase of \textcolor{red}{0.67} dB on the deraining task and \textcolor{red}{1.65} dB on the dehazing task compared to InstructIR \cite{conde2024high-InstructIR}.

\subsection{Results Comparisons on Five Tasks}
To further validate the method's effectiveness in addressing a broader range of tasks, building upon the recent research conducted by IDR \cite{all-in-one_IDR_zhang2023ingredient} and InstructIR \cite{conde2024high-InstructIR}, we extend our investigation into the efficacy of DFPIR by conducting experiments across five restoration tasks: dehazing, deraining, denoising, deblurring, and low-light image enhancement. 
To achieve this, we train a comprehensive DFPIR on combined datasets compiled for five distinct tasks. These encompass datasets from the previously mentioned three-task scenario, alongside additional datasets: GoPro \cite{nah2017deep} for motion deblurring and LOL \cite{wei2018deep} for low-light image enhancement.  
Tab. \ref{all-in-one-5T} demonstrates that DFPIR achieves a \textcolor{red}{1.09} dB improvement compared to the recent leading method InstructIR \cite{conde2024high-InstructIR}, when averaged across five restoration tasks. 
Additionally, we compare our method to general image restoration models that were trained in the same All-in-One setting. Notably, our method surpasses
Restormer\cite{IM_zamir2022restormer} and NAFNet \cite{IM_NAFNetchen2022simple}  on average PSNR by \textcolor{red}{3.04} dB and \textcolor{red}{2.88} dB, respectively, validating the effectiveness of our approach in handling multiple degradations.

\subsection{Visual Results}
\begin{figure}[t]
	\setlength{\abovecaptionskip}{0.cm}
	\centering
	\includegraphics[width=1\linewidth]{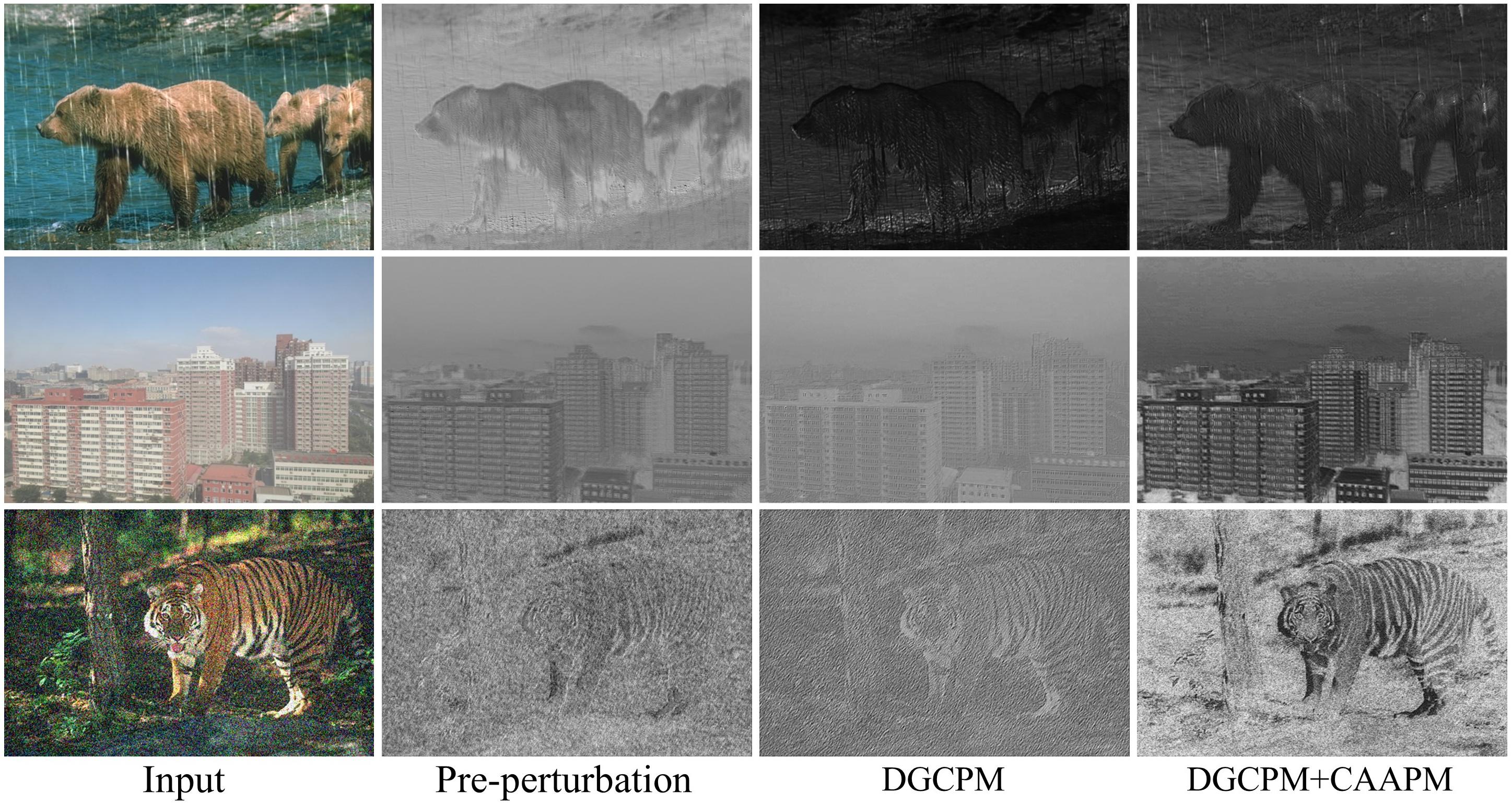} 
	\caption{Feature visualization. DGCPM extracts the intrinsic features of the image while suppressing degradation characteristics. DGCPM + CAAPM enhances intrinsic features while further reducing the impact of degradation. Zoom in for better view.}
	\label{feature_visual}
\end{figure}

\begin{figure}[htbp]
	\setlength{\abovecaptionskip}{0.cm}
	\centering
	\includegraphics[width=1\linewidth]{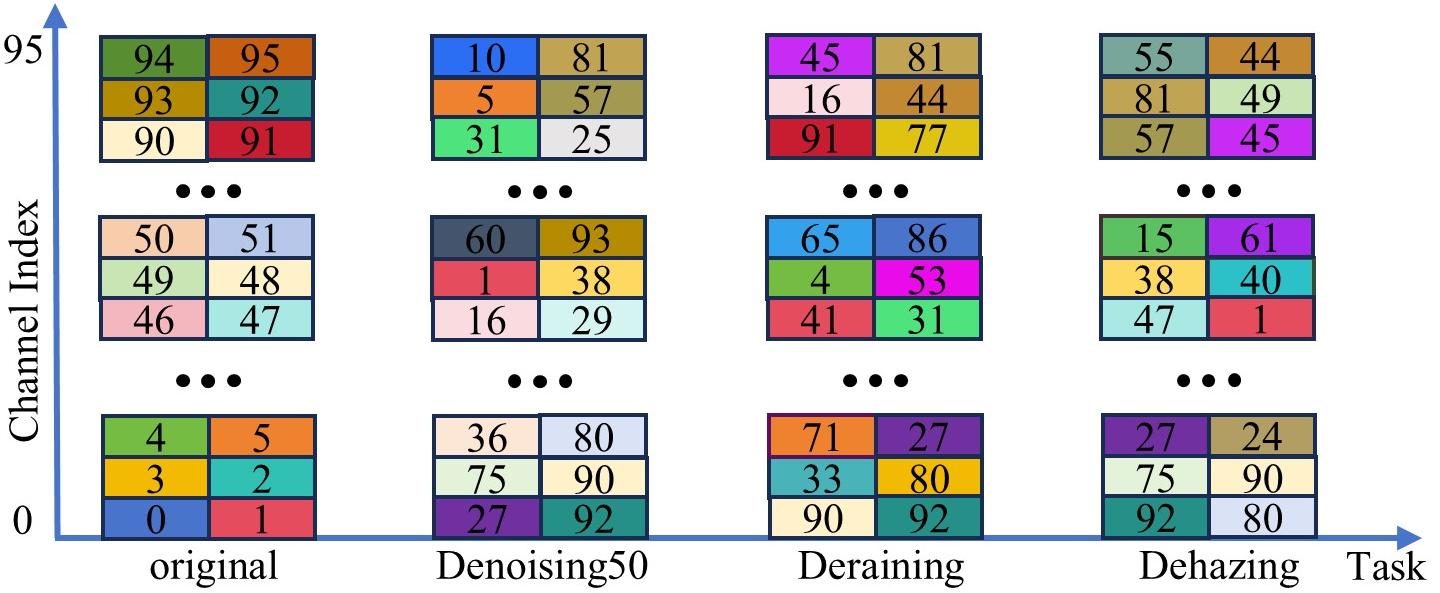} 
	\caption{Channel shuffle visualization. After channel shuffling, the feature channel order for each task has changed compared to the original order.}
	\label{channel_index}
\end{figure}

Visual examples illustrating our results for dehazing, deraining, and denoising are provided in Fig.\ref{visual result-3t}. Compared to PromptIR \cite{potlapalli2023promptir} and InstructIR \cite{conde2024high-InstructIR}, 
our method is more effective in challenging dehazing scenarios. Additionally, our model's deraining results are closer to the ground-truth images. Furthermore, in image denoising, our model recovers more details from heavily degraded noisy inputs. 
We visualized the results of channel shuffling, as shown in Fig.\ref{channel_index}. After channel shuffling, the channel order for each task changed accordingly, which validates the effectiveness of our proposed channel shuffling strategy. Additionally, we also visualized the feature perturbations before and after for the ${{\textbf{F}}_1}$ layer, as shown in Fig.\ref{feature_visual}. As can be seen, after channel dimension perturbation (DGCPM), the network extracts the image's inherent features from multi-degradation scenarios, reducing the impact of degradation-specific features. 
With the addition of perturbation in the attention dimension (DGCPM+CAAPM), the intrinsic detail features of the image are further enhanced, while the degradation features are more effectively suppressed. This demonstrates the efficacy of our proposed channel and attention perturbation strategy.

\subsection{Ablations Studies}

We conduct several ablation experiments to demonstrate the effectiveness of our proposed degradation-guided perurbation block. We report the results of training an all-in-one model on combined datasets from three restoration tasks. \textbf{More detailed ablation experiments can be found in the supplementary materials.}

\textbf{Impact of key components.} As illustrated in the Tab. \ref{ablation}(a), using channel attention (Method (a) ) directly improves by 0.36 dB compared to the baseline \cite{IM_zamir2022restormer}, but it is 0.15 dB lower than channel shuffle (Method (c) ). This also validates the effectiveness of the channel shuffle strategy we proposed. 
Channel shuffle preserves inherent image features with degradation info but offers limited reduction in cross-degradation interference. By applying attention-wise perturbation, restoration quality is significantly enhanced (DGCPM+CAAPM). The average PSNR increased from 32.49 to 32.88, reflecting an improvement of 0.39 dB. However, Method (b) (CA+CAAPM) results in a lower performance than DFPIR, indicating that the perturbations in channel and attention dimensions produce a synergistic enhancement effect.

\textbf{Impact of parameter $ \gamma $.}  As illustrated in the Tab. \ref{ablation}(b), if the perturbation in the attention dimension is too high ($ \gamma =0.5 $) or absent ($ \gamma =1.0 $), the performance is not optimal. This is because excessive perturbation, while reducing the interference between images with different degradations, increases information loss, leading to suboptimal performance. Similarly, if the perturbation is too small, the interference between tasks becomes more significant, resulting in suboptimal performance as well.

\begin{table}[t]
	\centering
	\caption{Ablation study results on three tasks. Average PSNR (dB, ↑) and SSIM (↑) metrics are reported on the full RGB images.The CA stands for Channel Attention.}
	\label{ablation}
	\begin{minipage}{0.27\textwidth}
		\centering
		\caption*{(a) Impact of key components. }
		\fontsize{8}{12.6}\selectfont
		\setlength{\tabcolsep}{1.2pt} 
		\begin{tabular}{c|ccc|c}
			\hline
			Method&CA & DGCPM & CAAPM & PSNR/SSIM  \\
			\hline
			baseline &$\times$     & $\times$  & $\times$ & 31.98 / 0.911  \\
			(a)      & \checkmark  & $\times$  & $\times$ & 32.34 /  0.914  \\
			(b)      & \checkmark  & $\times$  & \checkmark & 32.65 /  0.917  \\
			(c)      & $\times$    & \checkmark  & $\times$ & 32.49 /  0.910  \\
			\hline
			DFPIR &$\times$ &\checkmark &\checkmark  & 32.88 / 0.919 \\	
			\hline
		\end{tabular}
	\end{minipage}
		\hspace*{0.010\textwidth} 
	\hfill
	\begin{minipage}{0.18\textwidth}
		\centering
		\caption*{(b) Impact of $ \gamma $. }
		\fontsize{7.8}{12.6}\selectfont
		\setlength{\tabcolsep}{5pt} 
		\begin{tabular}{c| c}
			\hline
			$ \gamma $  & PSNR / SSIM  \\
			\hline
			 0.5 & 32.67 / 0.913 \\
			 0.7 & 32.81 / 0.919 \\
			 0.8 & 32.84 / 0.919 \\
			 0.9 & 32.88 / 0.919 \\	
			 1.0 & 32.83 / 0.919 \\		
			\hline
		\end{tabular}
	\end{minipage}
	
\end{table}

\section{Conclusion}

In this paper, we present a novel all-in-one image restoration framework, DFPIR, which introduces Degradation-aware Feature Perturbations (DFP) to adjust the feature space in alignment with a unified parameter space. Our approach incorporates both channel-wise and attention-wise perturbations, dynamically guided by degradation type prompts. Specifically, channel-wise perturbations are achieved by shuffling channels in a high-dimensional space, while attention-wise perturbations are implemented through selective masking in the attention space.
To effectively realize these operations, we design a Degradation-Guided Perturbation Block (DGPB), which integrates channel shuffling and attention masking, strategically placed between the encoding and decoding stages of the encoder-decoder architecture.
The proposed DGPB demonstrates its efficacy in enhancing comprehensive image restoration when integrated into a state-of-the-art model, yielding notable improvements in the all-in-one restoration setting. 

\vspace{-2.5pt}  
\section{Acknowledgment}
This work was supported in part by the National Natural Science Foundation of China under Grant 62171304 and partly by the Natural Science Foundation of Sichuan Province under Grant 2024NSFSC1423, Cooperation Science and Technology Project of Sichuan University and Dazhou City under Grant 2022CDDZ-09, the TCL Science and Technology Innovation Fund under grant 25JZH008, and the Young Faculty Technology Innovation Capacity Enhancement Program of Sichuan University under Grant 2024SCUQJTX025.

\newpage
{
    \small
    \bibliographystyle{ieeenat_fullname}
    \bibliography{main}
}


\end{document}